\documentclass[letterpaper]{article}

\usepackage{natbib,alifeconf}  
\usepackage{amsmath}
\usepackage{url,hyperref}
\usepackage{booktabs}

\usepackage{subcaption}
\usepackage{float}
\usepackage[export]{adjustbox}
\usepackage[table]{xcolor}
\usepackage{multirow}
\usepackage{cleveref}

\newcommand{\abs}[1]{\left|#1\right|} 

\title{ARC-NCA: Towards Developmental Solutions to the \\Abstraction and Reasoning Corpus}

\author{
    Etienne Guichard$^{1}$,
    Felix Reimers$^{1}$, \and
    Mia Kvalsund$^{2}$, \and
    Mikkel Lepper$\o$d$^{3}$, \and
    Stefano Nichele$^{1,4}$\\
    \mbox{}\\
    $^1$$\O$stfold University College, Halden, Norway \\
    $^2$University of Oslo, Oslo, Norway \\
    $^3$Simula Research Laboratory, Oslo, Norway \\
    $^4$Oslo Metropolitan University, Oslo, Norway \\
    stefano.nichele@hiof.no
} 

\begin{document}

\maketitle
\begin{abstract}
The \textit{Abstraction and Reasoning Corpus (ARC)}, later renamed ARC-AGI, poses a fundamental challenge in artificial general intelligence (AGI), requiring solutions that exhibit robust abstraction and reasoning capabilities across diverse tasks, while only few (with median count of three) correct examples are presented. While ARC-AGI remains very challenging for artificial intelligence systems, it is rather easy for humans. This paper introduces ARC-NCA, a developmental approach leveraging standard Neural Cellular Automata (NCA) and NCA enhanced with hidden memories (EngramNCA) to tackle the ARC-AGI benchmark. NCAs are employed for their inherent ability to simulate complex dynamics and emergent patterns, mimicking developmental processes observed in biological systems. Developmental solutions may offer a promising avenue for enhancing AI's problem-solving capabilites beyond mere training data extrapolation. ARC-NCA demonstrates how integrating developmental principles into computational models can foster adaptive reasoning and abstraction. We show that our ARC-NCA proof-of-concept results may be comparable to, and sometimes surpass, that of ChatGPT 4.5, at a fraction of the cost. 

\end{abstract}

\textbf{Data/Code}: \url{https://github.com/etimush/ARC_NCA}\\

\textbf{Videos}: \url{https://etimush.github.io/ARC-NCA-Videos/}\\

Submission type: \textbf{Full Paper}\\

\section{Introduction}
\begin{figure}[t!]
\centering
\includegraphics[width=0.45\textwidth]{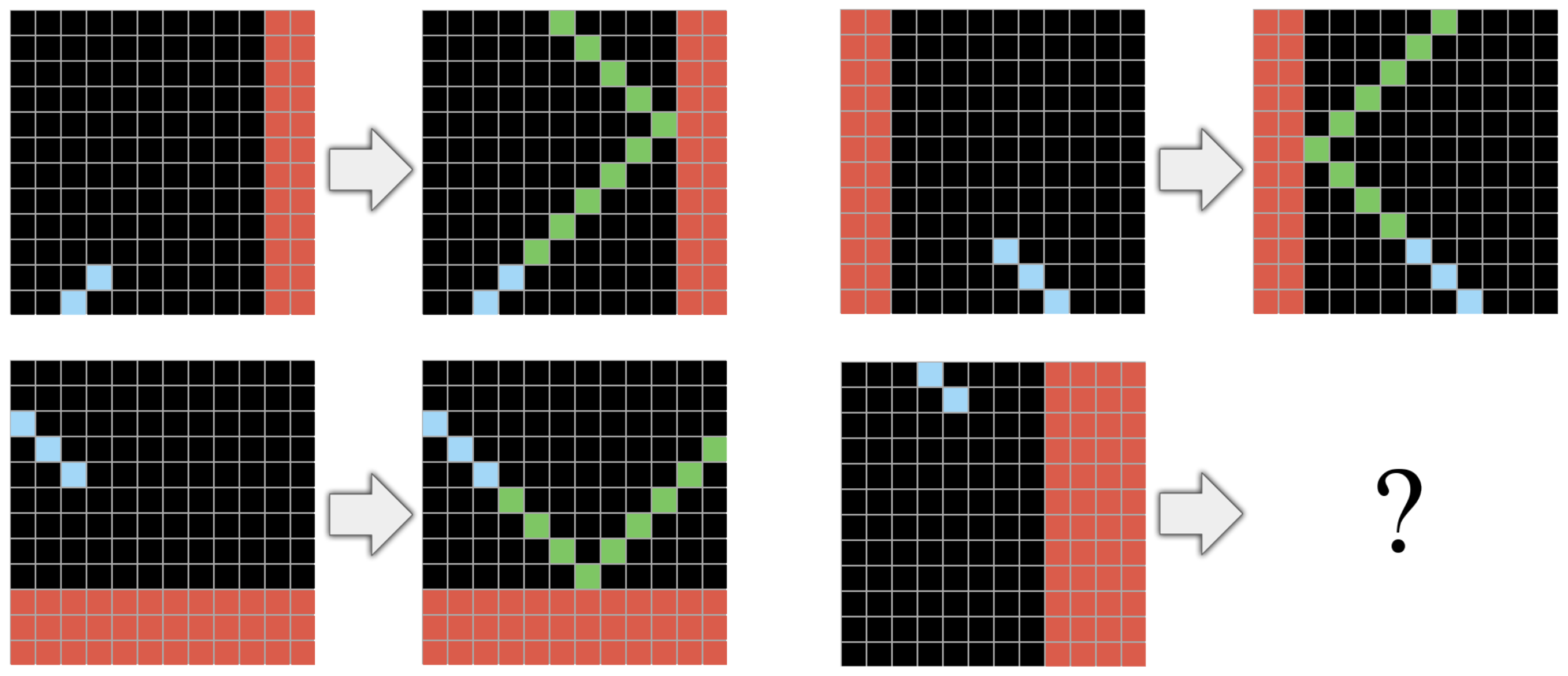}
\caption{Example ARC task, adapted from \citep{chollet2019measure}.}
\label{fig:ARC-AGI-example}
\end{figure}

Progress towards artificial general intelligence (AGI) necessitates benchmarks that rigorously assess an agent's capacity for abstraction, generalization, and reasoning. The Abstraction and Reasoning Corpus (ARC), introduced by \citep{chollet2019measure}, is one of such benchmarks. It comprises a collection of visual pattern transformation tasks, each defined by a few input-output examples, challenging AI models to infer the underlying transformation rules and apply them to novel instances. Test pairs consist of two components: an "input grid," which is a rectangular array of cells with varying dimensions (up to 30 rows by 30 columns), where each cell holds one of ten distinct "values," and an "output grid," which can be entirely derived from the attributes and structure of the input grid. One of such tasks is depicted in Figure \ref{fig:ARC-AGI-example}. The purpose is to examine the example pairs to grasp the nature of the problem and utilize this understanding to produce the corresponding output grid for each given test input. Two attempts can be performed for each input grid. Each task is handcrafted by human designers with a unique logical structure, making it very difficult to prepare for each task in advance.
Such emphasis on few-shot learning and the necessity for broad generalization make ARC-AGI particularly demanding for current AI systems.

In contrast, humans excel at these tasks, leveraging innate cognitive abilities to discern patterns and apply abstract reasoning with minimal examples. This disparity underscores a fundamental gap in current AI methods and highlights the need for novel approaches.

\begin{figure*}[t!]
\centering
\includegraphics[width=0.75\textwidth]{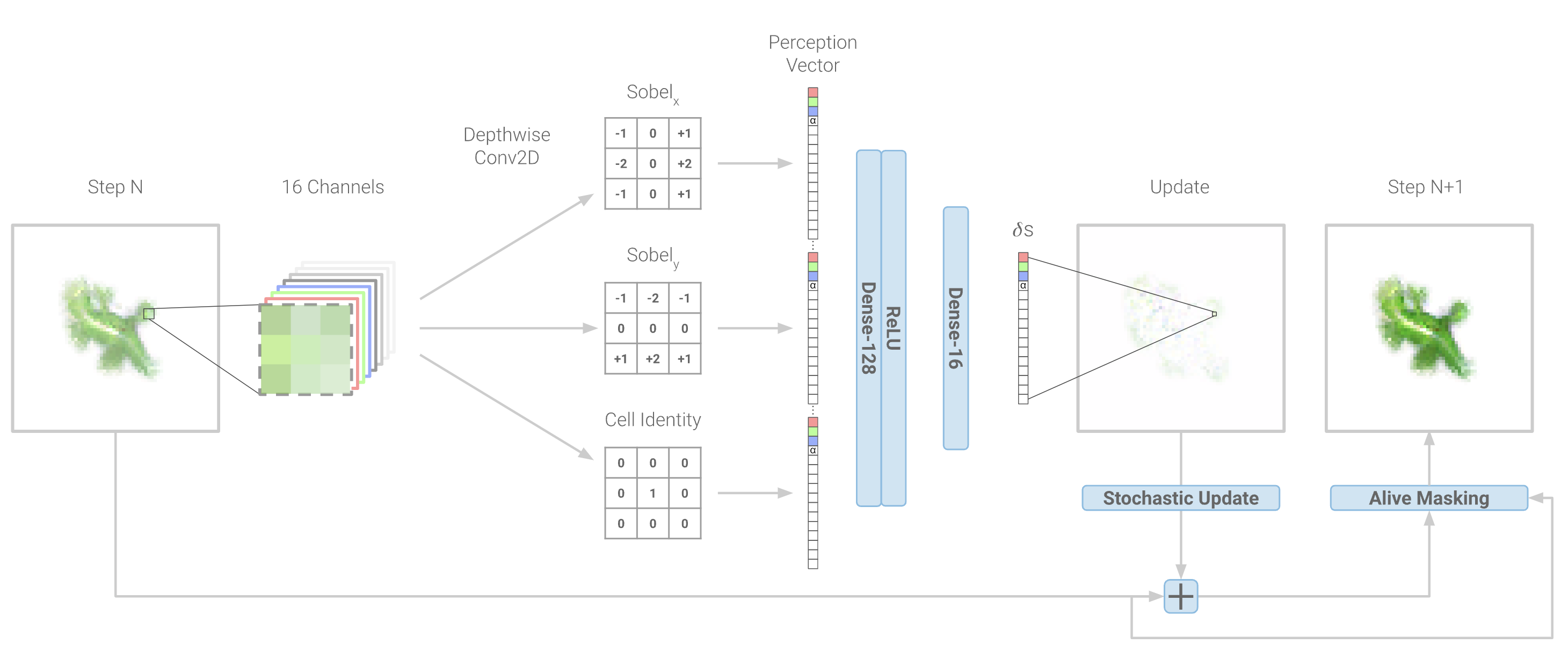}
\caption{Diagram depicting one pass of the Growing NCA update step and its neural network model. Adapted from \citep{mordvintsev2020growing}.}
\label{fig:GNCA-model}
\end{figure*}

One promising avenue lies in the realm of developmental computation, inspired by the processes observed in living systems. Neural Cellular Automata (NCA) \citep{gilpin2019cellular, mordvintsev2020growing, nichele2017neat}, exemplify this approach. NCAs are computational models where each cell on a lattice updates its state based on local interactions governed by neural networks, leading to the emergence of complex global patterns. As such, NCAs have been used as models of biological morphogenesis \citep{randazzo2023biomaker, stovold2023neural, pontes2022single, sudhakaran2021growing}, where local cellular interactions give rise to organized structures during development (such as bodies and brains). Furthermore, biological brains employ cognitive mechanisms that may mirror developmental processes to facilitate reasoning, abstraction, and problem-solving through dynamic, iterative, and self-organizing processes. Examples include iterative refinement of mental schemas through interactions with the environment \citep{mcvee2005schema, neumann2018use}, hierarchical structuring to break down tasks in sub-tasks \citep{botvinick2009hierarchically, meunier2009hierarchical}, and predictive modeling to anticipate outcomes and proactively adjust solutions \citep{friston2003learning, seth2014cybernetic, millidge2021predictive}. The hypothesis tested in this work is whether the developmental nature of NCAs makes them particularly suited for tasks like those in the ARC-AGI benchmark.

In the last years, most approaches for ARC-AGI relied on discrete program search, a brute force methodology. Recently, Large Language Models (LLMs) have been utilized in different ways, including for optimizing domain-specific languages \citep{chollet2024arc}. Further, LLMs have been used for program synthesis with the intention of generating programs in general-purpose languages, e.g., Python, that attempt to solve the task at hand. Test-time training, also known as inference-time fine-tuning, has been rather popular in the last few months to allow inference-time adaptations based on unseen test samples. Often, hybrid approaches, including program synthesis and transductions, i.e., direct prompting an LLM, have been combined. However, solving the ARC is still an open problem and the solution might still lie in uncharted areas of model selection.

In this paper, we introduce ARC-NCA, a novel approach that leverages the developmental dynamics of standard Neural Cellular Automata \citep{mordvintsev2020growing} and an enhanced variant with hidden memory states, termed EngramNCA \citep{guichard2025engramnca}, to tackle the ARC-AGI benchmark. To the best of our knowledge, this is the first time NCAs are used for the 2D ARC-AGI benchmark. EngramNCA is chosen, in addition to standard NCAs, because it relies on mechanisms for learning low-level morphologies and manipulations first, and then a regulation mechanism for deciding when and where such primitives should be activated and propagated, which is considered a suitable mechanism for abstraction and reasoning tasks. By emulating the principles of biological development and cognitive development, our models aim to capture essential aspects of human-like abstraction and reasoning. Our ARC-NCA approach may be considered a program synthesis approach, where a custom NCA (a "program") is generated for the task at hand with a fine-tuning process akin to test-time training. Our proof-of-concept demonstrates that ARC-NCA may reach performances comparable, and sometimes superior, to existing models (including ChatGPT 4.5, see results and discussion section for details), but with significantly reduced computational resources.
We hope that our work will spark a renewed interest within the artificial life community for radically new approaches to abstraction and reasoning. 



\section{Related Works}

The application of cellular automata (CA) models, and morphogenetic models in general \citep{wolfram1997new}, to the ARC-AGI benchmark \citep{chollet2019measure} remains an underexplored area. However, several developments in CA research suggest potential avenues for applying CA methodologies to ARC tasks.
In particular, one architectural choice that opens up opportunities for learning CA rules is Neural Cellular Automata \citep{gilpin2019cellular, mordvintsev2020growing, nichele2017neat}, where a neural network replaces more traditional CA lookup tables. NCA was proposed as a possible embodied controller by \citep{variengien2021towards}, where an NCA was connected to a reinforcement learning environment in a closed loop, thus demonstrating a self-organising "brain". Another interesting line of research is aiming at critical NCAs \citep{pontescritical, Guichard2024}, i.e., CA models operating at the edge-of-chaos \citep{langton1990computation}, which could be a powerful pre-training strategy. A NCA for image manipulation, named Vision Transformer Cellular Automata (ViTCA), is proposed in \citep{tesfaldet2022attention}, where attention heads are included in the model, inspired by the transformer architecture \citep{vaswani2017attention}. \citep{Reimers2023} proposes a variation with local-self attention, while \citep{kvalsund2024sensor} presents an evolved attention-like mechanism. In general, transformers can learn elementary CA rules \citep{burtsev2024learning}, opening up interesting opportunities of potentially combining CA and LLMs for ARC-AGI in the future. 
A work using an evolutionary approach is \citep{fischer2020solving}, where grammatical evolution is employed for optimizing expressions in a domain-specific language for incremental image transformations. An accelerated JAX implementation of CA, including NCA, is proposed in \citep{faldor2024cax}, where they also attempt to employ a 1D-NCA for solving the much simpler 1D-ARC dataset \citep{xu2023llms}, an unofficial simplified adaptation of ARC-AGI composed of 1-dimensional rows of pixels, which significantly reduces task complexity. For a recent report on popular attempts at solving the ARC-AGI challenge, including program synthesis methods with deep learning techniques, please refer to \citep{chollet2024arc}. Very recently, in April 2025, OpenAI has announced that their most powerful models at that time, named o3 and o4 mini (two reasoning models using support tokens for planning and summoning internal tokens to run Python code as part of their reasoning, before providing an answer), achieved promising scores in ARC-AGI \citep{ARCprizeo3, ARCprizeanalizingo3}. Specifically, o3-low scored 41\%, o3-medium 53\%, o4-mini-low 21\%, and o4-mini-medium 41\%, all on the semi-private evaluation set. Additionally, two o3 versions tested with high compute resources (namely using 6 and 1024 independent inference samples) scored 75.7\% and 87.5\%, using 33.5 million and 5.7 billion tokens. The reported cost for the version with 6 inference samples was 201 USD per sample, while the version with 1024 was 172x more expensive. This staggering cost might be significantly reduced by alternative architectures.

\begin{figure*}[t!]
\centering
\includegraphics[width=0.75\textwidth]{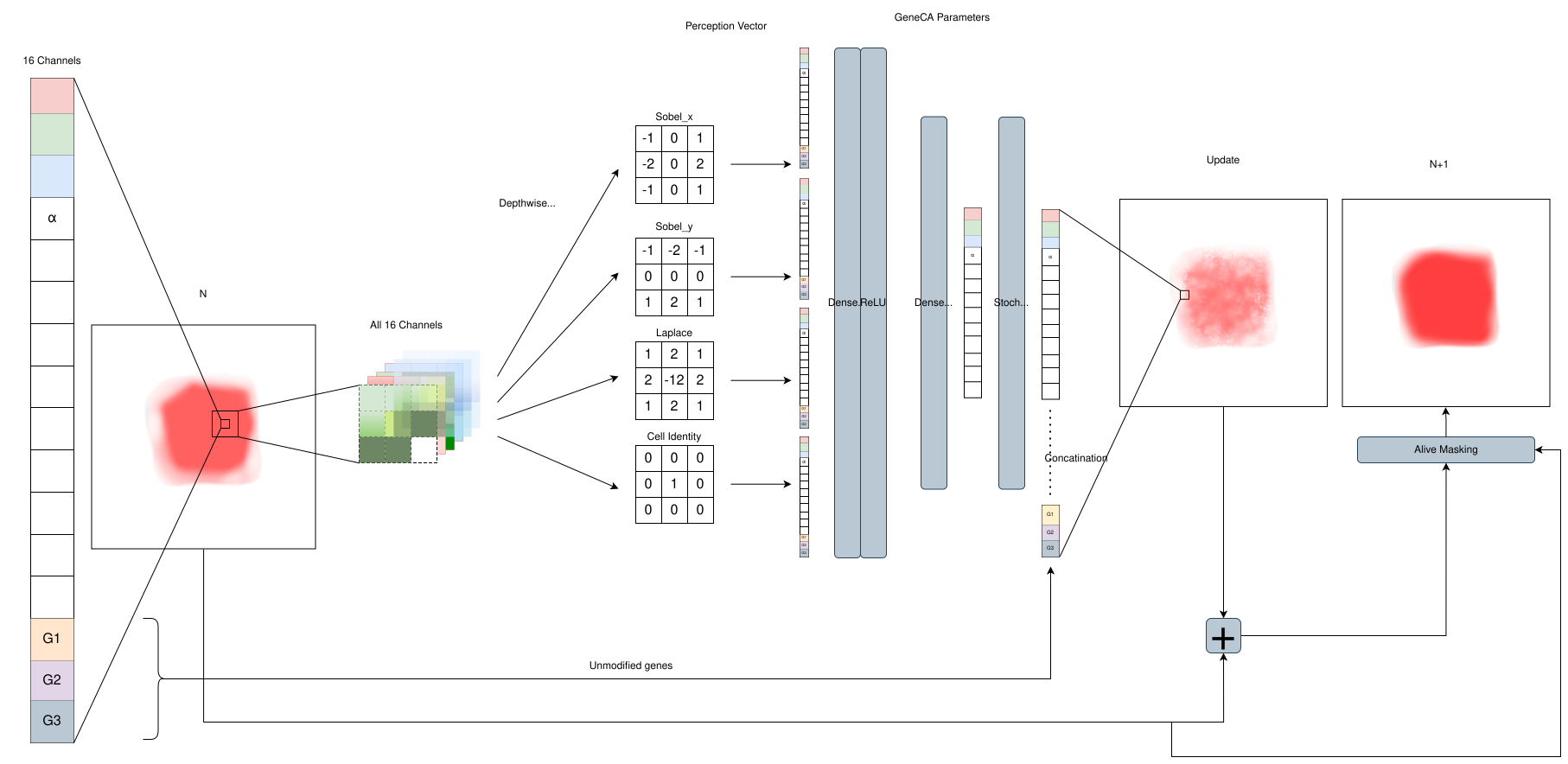}
\caption{Diagram depicting one pass of the EngramNCA GeneCA update step and its neural network model. Adapted from \citep{guichard2025engramnca}.}
\label{fig:GeneCA-model}
\end{figure*}

\begin{figure*}[t!]
\centering
\includegraphics[width=0.75\textwidth]{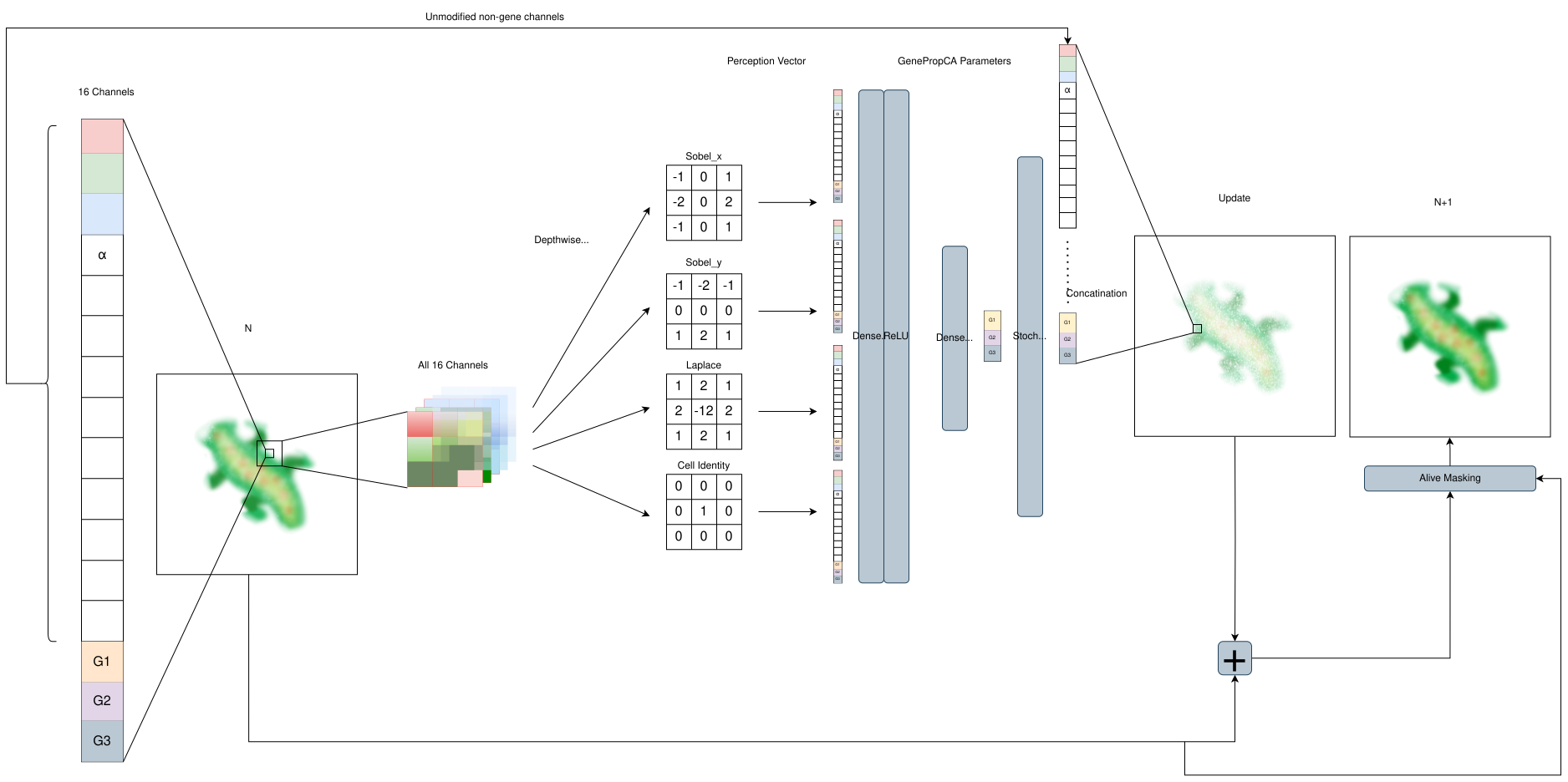}
\caption{Diagram depicting one pass of the EngramNCA GenePropCA update step and its neural network model. Adapted from \citep{guichard2025engramnca}.}
\label{fig:GenePropCA-model}
\end{figure*}

\section{Models and Methods}
This section details the models used in obtaining developmental solutions to the \textit{Abstraction and Reasoning Corpus}. We chiefly explore NCA models and their derivatives in the form of classic NCA and EngramNCA (and modifications to EngramNCA).
\subsection{NCA models}
We choose to test the Growing NCA as presented by \citep{mordvintsev2020growing}, along with four versions of EngramNCA presented in \citep{guichard2025engramnca}: EngramNCA v1, an unmodified version of EngramNCA, EngramNCA v2,  v3, and v4, modified versions of EngramNCA with ARC-specific augmentations.  

We believe the standard NCA model needs no detailed introduction. In short, it is implemented as a differentiable neural network embedded in a cellular automaton framework, where each cell maintains a continuous state vector updated through convolutional neural networks (CNNs) with learned local update rules. The architecture is depicted in Figure \ref{fig:GNCA-model}. However, EngramNCA is a relatively recent model and thus warrants a brief introduction. Its NCA features dual-state cells with distinct public (interaction-based) and private (memory-based) states. The model is an ensemble that includes: GeneCA, an NCA which generates morphological patterns from a seed cell encoding genetic primitives (Figure \ref{fig:GeneCA-model}); GenePropCA,  an NCA which propagates and activates these genetic primitives across the cell network (Figure \ref{fig:GenePropCA-model}), similar to RNA-based communication \citep{shomrat2013automated}. EngramNCA is trained in two stages: first, GeneCA is trained to grow primitive morphologies containing immutable private memory encodings, using only publicly visible channels for coordination; then, GenePropCA is trained to modulate the private memory of cells without altering their visible states, enabling the transfer of genetic information across the grid. For details on the model see \citep{guichard2025engramnca}.

\begin{table}[h!]
\centering

\begin{tabular}{ |p{0.13\textwidth}|p{0.13\textwidth}|p{0.13\textwidth}|}
 \hline
 \multicolumn{3}{|c|}{\textbf{CA Architecture Details}} \\
 \hline
 \textbf{CA type}& \textbf{Augmentations} & \textbf{Channels, Hidden Size} \\
 \hline
 NCA   & None (standard NCA) &    50, 64\\
 \hline
 EngramNCA v1&  None (standard EngramNCA)  & 50, (32,32) \\
 \hline
 EngramNCA v2 &  Sensing  & 50, (32,32) \\
 \hline
 EngramNCA v3 & Sensing + Toroidal & 50, (32,32)\\
 \hline
 EngramNCA v4   &Sensing + Toroidal + Local vs Global & 50, (32,32)\\
 \hline
\end{tabular}
\caption{Architecture detail for all  CA variants. The different notations for NCA and EngramNCA on \textit{Channels, Hidden Size} are due to the split versus standard architecture between the two. 
}
 \label{tab:CAArchi}
\end{table}

Table \ref{tab:CAArchi} shows the different CA architectures. The augmentations are detailed in sections \nameref{sec: LvG}, \nameref{sec:TvNT}, and \nameref{sec:sense}. 

\subsection{From ARC to NCA Space}\label{sec:FromAtoN}
The ARC dataset mainly comprises $2D$ grids with integer values. Each grid can range from 1x1 to 30x30 in size, with values ranging between 0 and 9.

We address two major issues with transforming ARC grids into NCA-compatible ones:

\subsubsection{From $2D$ Integer grid to $3D$ real-valued lattice.}

NCA mainly operate on $3D$ lattices of dimensions $H,W,C$, where $H$ is the height, $W$ is the width, and $C$ is the number of channels, most commonly four channels for $RGB-\alpha$ values of an image, and an arbitrary amount of hidden channels. To transform the ARC grids into NCA lattices, we first assume two conditions:

\begin{itemize}
    \item Constant $\alpha$: all colors represented by the 10 integers have the same alpha value of 1
    \item Equal spacing: all 10 colors (0-9) are equally spaced apart in an HSL (hue, saturation, and lightness) color spectrum, starting with black for 0.
\end{itemize}

We then transform the ARC problems into an $RGB-\alpha$ lattice using an integer-to-HSL-to-$RGB-\alpha$ conversion equation:

\begin{align}
    h &= \frac{v}{n} \times 360 \quad \text{(Hue calculation)} \\
    l &= 0.5 \quad \text{(Fixed Lightness)} \\
    s &= 0.8 \quad \text{(Fixed Saturation)} \\
    C &= (1 - \abs {2l - 1}) \times s \times (v>0) \quad \text{(Chroma)} \\
    M &= (l - \frac{C}{2}) \times (v>0).
\end{align}

Here, $v$ is the integer value in that grid position, and $n$ is the total number of colors.

\begin{align}
    X &= c \times \left(1 - \abs{ \left(\frac{h}{60} \pmod 2\right) - 1 }\right) 
\end{align}

\begin{align*}
    R' &=
    \begin{cases}
        C & \text{if } 0^\circ \le h < 60^\circ \\
        X & \text{if } 60^\circ \le h < 120^\circ \\
        X & \text{if } 240^\circ \le h < 300^\circ \\
        C & \text{if } 300^\circ \le h < 360^\circ \\
        0 & \text{otherwise}
    \end{cases} \\
    G' &=
    \begin{cases}
        X & \text{if } 0^\circ \le h < 60^\circ \\
        C & \text{if } 60^\circ \le h < 120^\circ \\
        X & \text{if } 120^\circ \le h < 180^\circ \\
        X & \text{if } 180^\circ \le h < 240^\circ \\ 
        0 & \text{otherwise}
    \end{cases} \\
    B' &=
    \begin{cases}
        X & \text{if } 120^\circ \le h < 180^\circ \\
        C & \text{if } 180^\circ \le h < 240^\circ \\
        C & \text{if } 240^\circ \le h < 300^\circ \\
        X & \text{if } 300^\circ \le h < 360^\circ \\
        0 & \text{otherwise}
    \end{cases}
\end{align*}

\begin{align}
    R &= (R' + M) \times 255 \\
    G &= (G' + M) \times 255 \quad  \\
    B &= (B' + M) \times 255 \\
    \alpha &= 255 * (v>0)
\end{align}

We extend the channel dimension of the $RGB-\alpha$ lattice with a binary encoding for each pixel based on its color. We finally pad the channel dimension with ones to reach the desired number of hidden channels.

\begin{figure*}[t!]
\centering
\includegraphics[width=0.80\textwidth]{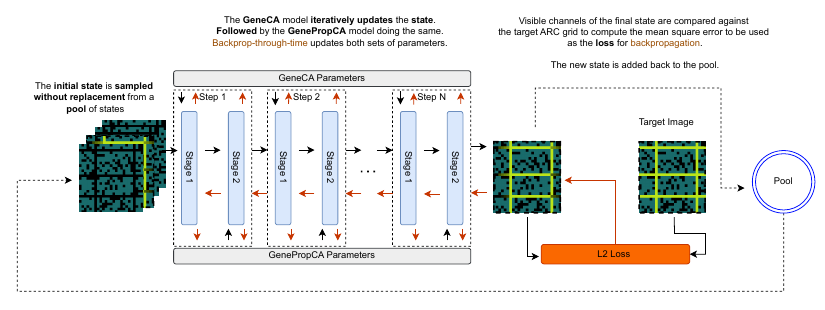}
\caption{One backpropagation step of training EngramNCA for solving ARC problems.}
\label{fig:ca_training}
\end{figure*}

\subsubsection{Dealing with changing grid sizes.} \label{sec:sizes}
Certain ARC problems contain solutions whose grid size differs from the input. This presents a pernicious problem in that NCAs cannot modify their grid size. To deal with this, we explore two methods:

\begin{itemize}
    \item Ignore the problematic grids: Remove them from the training procedure.
    \item Maximal size padding: Pad every problem to the maximal 30x30 grid size with a special padding value, one uniquely only found in the padded areas, and allow the NCA to modify the amount of padding.
\end{itemize}

Due to computational constraints, we choose to mainly focus on ignoring the problematic grids. However, \nameref{sec:Further} details the experiments done with Maximal size padding. All results will be reported on the 262 problems that do not require resizing. 

\subsection{ARC Specific Augmentations}

The ARC dataset provides some specific challenges that NCA can have difficulties dealing with, one such challenge was discussed in \nameref{sec:FromAtoN}. However, we also identified three other challenges: 

\subsubsection{Toroidal versus Non-Toroidal Problems}\label{sec:TvNT}- In general, NCA operates on a toroidal lattice. While this is desirable for tasks such as growing morphologies, as it means the morphology is positionally invariant, it causes issues in ARC-AGI problems where absolute positions and grid edges are a necessary part of the reasoning. Disabling this behavior is also not a reasonable option, as some ARC-AGI problems become easier to solve if information propagates toroidally. 

We remedy this in EngramNCA v3 and EngramNCA v4 in two ways: by splitting the functionality of the GeneCA and GenePropCA. The former acts on a non-toroidal lattice, while the latter acts on a toroidal lattice, and by giving each cell channel-wise local self-attention. 

The hypothesis is that by splitting the functionality and imbuing it with attention, the EngramNCA might be able to choose whether or not it exhibits toroidal functionality.

\subsubsection{Local versus Global Solutions}\label{sec: LvG}-
Another problem comes in the form of whether the NCA should focus on global or local information when solving ARC problems (or a mix of the two). This should, in theory, not be a problem. However, we qualitatively observe that some problems struggle with fine-grained local information. 

We introduce a patch training scheme to force the NCA to focus on local information. This scheme works on the same principle as the standard way of NCA training, with the key difference being that the NCA is trained on- and loss is accumulated over- 3x3 patches of the grid, instead of the entire grid at once. Since this is an augmentation to NCA training, it becomes more costly to train the NCA, thus, we choose to only use this augmentation in EngramNCA v4.

\subsubsection{Inappropriate Sensing}- \label{sec:sense}
Due to NCA's initial applications being the simulation of the growth of organisms, the sensing mechanisms somewhat mimic biological cells' chemosensing mechanisms, in the form of gradient sensing kernels. While a helpful analogy, this might present a fundamental limitation for the purposes of ARC. To combat this, we choose to augment EngramNCA v3 and EngramNCA v4 with fully learnable sensing filters in place of the Sobel and Laplacian filters. The number of filters was chosen to match that of the standard EngramNCA.

\section{Training}
\subsection{Determining the Quality of Solutions}

During training, the NCA effectively produces an image. We ostensibly do not consider the developmental steps the NCA takes to reach the final solution. Thus, we take the loss to be $MSEPixelWiseLoss$ as in \citep{mordvintsev2020growing}. 

To determine whether a problem is solved, we look at the mean pixel error across the generated NCA. An evaluation loss of $log(MSEPixelWiseLoss) \leq -7$, where this loss was evenly distributed among pixels, was experimentally determined to produce exact solutions to the ARC problems.

\subsection{Model Training}

We choose to solve ARC via test-time training. As stated by \citep{chollet2019measure, chollet2024arc}, program generators must be able to learn from new information. We take this to mean that our program generator, the system that trains NCAs, can train a new CA per problem. For every problem, we train a new CA from scratch on the 2-3 training examples and evaluate its performance on the unseen sample. All our experiments are run on the ARC-AGI public evaluation set.

Figure \ref{fig:ca_training} shows one training iteration of the training procedure for EngramNCA versions. The training procedure mirrors that of \citep{guichard2025engramnca} with one key modification. Due to training both the GeneCA and GenePropCA from scratch for each problem, the GeneCA weights are not frozen, and both sets of weights are co-optimized. The standard NCA was instead trained with the same procedure shown in \citep{mordvintsev2020growing}.

\begin{equation}
\resizebox{0.43\textwidth}{!}{$PixelWiseMSE = \frac{1}{H \times W \times C} \sum_{i=0}^{H} \sum_{j=0}^{W} \sum_{k=0  }^{C} \left( I(i, j, k) - \hat{I}(i, j, k) \right)^2$}
\end{equation}

Where $H$, $W$, $C$ are the dimensions of the image, $I$ is the reference image, and $\hat{I}$ is the final state of the NCA.

We use AdamW as the optimizer, with a learning rate (LR) of $1e-3$. For each problem, the CA are trained for 3000 iterations, with a 66\%  reduction in LR at 2000 iterations.

\section{Results}
\subsection{General Results}
In this section, we present the results of each CA in the form of Mean $log(loss)$ and the CA solve rate. The same results were obtained for the union of different CA. As a reminder to the reader, two answers may be submitted when solving ARC; thus, by taking the union (each model produces one output) we still produce a valid submission.
\\
\begin{table}[h!]
\centering
\begin{tabular}{ |p{0.15\textwidth}||p{0.13\textwidth}|p{0.13\textwidth}|}
 \hline
 \multicolumn{3}{|c|}{CA Results} \\
 \hline
 Model& Mean $log(loss)$ &Solve Rate \\
 \hline
 NCA   & -4.31    &10.7\%\\
 EngramNCA v1&  \cellcolor{red!25} -3.63  & \cellcolor{red!25}6.5\% \\
 EngramNCA v2$\star$&   -4.03  & 9.2\% \\
 EngramNCA v3 & \cellcolor{green!25}-4.35 & \cellcolor{green!25}12.9\%\\
 EngramNCA v4   &-4.20 & 10.3\%\\
 Chat GPT 4.5$\star\star$ & N/A & 10.3\% \\
 \hline
 
\end{tabular}
\caption{Mean $log(loss)$ and solve rate for all four CA variations. The best results are highlighted in green, and the worst results are highlighted in red. $\star$Due to space constraints in the paper, and the fact that the results are very similar to EngramNCA v4, we omit EngramNCA v2 from many of the result discussions, including the unions of models. $\star\star$The results for Chat GPT 4.5 are taken from the ARC-AGI leaderboard \citep{ARCprizeLeader}. Note that such results were obtained on the ARC-AGI private evaluation set, instead of the public evaluation set as for our results.}
 \label{tab:GeneralResults}
\end{table}

Table \ref{tab:GeneralResults} shows the mean $loss(log)$ and solve rate for each CA. EngramNCA v3 performs best in both categories with a near 13\% solve rate. In contrast, EngramNCA v1 performs the worst in both metrics, with a solve rate of 6.5\%.

\begin{table}[h!]
\centering
\begin{tabular}{ |p{0.2\textwidth}||p{0.2\textwidth}|}
 \hline
 \multicolumn{2}{|c|}{Modle Cost/Task} \\
 \hline
 Model & Cost (\$/Task) \\
 \hline
 NCA   & $\approx$\cellcolor{green!25} 3$e$-4    \\
 EngramNCA v1&  $\approx$\cellcolor{green!25} 3$e$-4   \\
 EngramNCA v2&   $\approx$ 4$e$-4   \\
 EngramNCA v3 &  $\approx$ 4$e$-4 \\
 EngramNCA v4   & $\approx$ 5$e$-4 \\
 Chat GPT 4.5$\star$ & 0.29 \\
 \hline
 
\end{tabular}
\caption{Cost per task for each model. We calculate the estimated cost of our models on an NVIDIA RTX 4070 Ti by taking the average time and power usage (W) to train a task for each model and multiplying it by the cost/kWh in our area (\$0.37/kWh). Chat GPT 4.5 cost taken from \citep{ARCprizeLeader}. }
 \label{tab:cost}
\end{table}

Table \ref{tab:cost} shows the cost comparison between the CA models we experimented with and Chat GPT 4.5. We chose to compare to Chat GPT 4.5 as it has solve rates similar to ours and is one of the most popular LLMs. At roughly the same performance, we see a 1000x decrease in cost.

\begin{table}[h!]
\centering

\begin{tabular}{ |p{0.17\textwidth}||p{0.11\textwidth}|p{0.09\textwidth}|}
 \hline
 \multicolumn{3}{|c|}{CA Union Results} \\
 \hline
 CA& Mean $log(loss)$ &Solve Rate \\
 \hline
 NCA $\cup$ v1  & -3.97    & 13.7\%\\
 
 NCA $\cup$ v3& \cellcolor{green!25}  -4.32  & 14.8\% \\
 NCA $\cup$ v4 & -4.25 & 13.7\%\\
 v1 $\cup$ v3   &-3.98 & \cellcolor{green!25}15.3\%\\
 v3 $\cup$ v4   &-4.27 & 14.8\%\\
 v1 $\cup$ v4   &\cellcolor{red!25}-3.92 & \cellcolor{red!25}12.5\%\\
NCA $\cup$ v1 $\cup$ v3 $\cup$ v4& \cellcolor{gray!25}  -4.12  & \cellcolor{gray!25}17.6\% \\
 
 \hline
\end{tabular}
\caption{Mean $log(loss)$ and solve rate for unions.}
 \label{tab:UnionResults}
\end{table}

Table \ref{tab:UnionResults} shows the mean $log(loss)$ and solve rate for six unions of the CA. In this case, the union of EngramNCa v1 and EngramNCA v3 performs best for the solve rate, with a 15.3\%  solve rate. Effectively, half of the EngramNCA v1 solutions were not found in EngramNCA v3. All unions perform roughly equal or better than the single best model, indicating that all models have some non-overlapping problems they can solve. NCA and EngramNCA v3 performed best for mean $loss(log)$, which is to be expected as they both had the lowest mean losses. EngramNCA v1 and EngramNCA v4 performed worst in both categories. 

\subsection{Solved Problems}

In this section, we highlight one solved problem per CA type to show the developmental steps the CA models take to solve ARC problems. More video examples can be found \hyperlink{https://etimush.github.io/ARC-NCA-Videos/}{here}.

\begin{figure}[t!]
\centering
\begin{subfigure}[b]{0.47\textwidth}
\includegraphics[width=\textwidth]{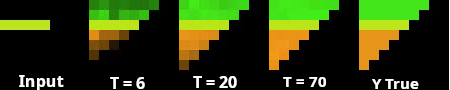}
\caption{Example  solution generated by the standard NCA.}
\label{fig:NCAsol}
\end{subfigure}

\begin{subfigure}[b]{0.47\textwidth}
\centering
\includegraphics[width=\textwidth]{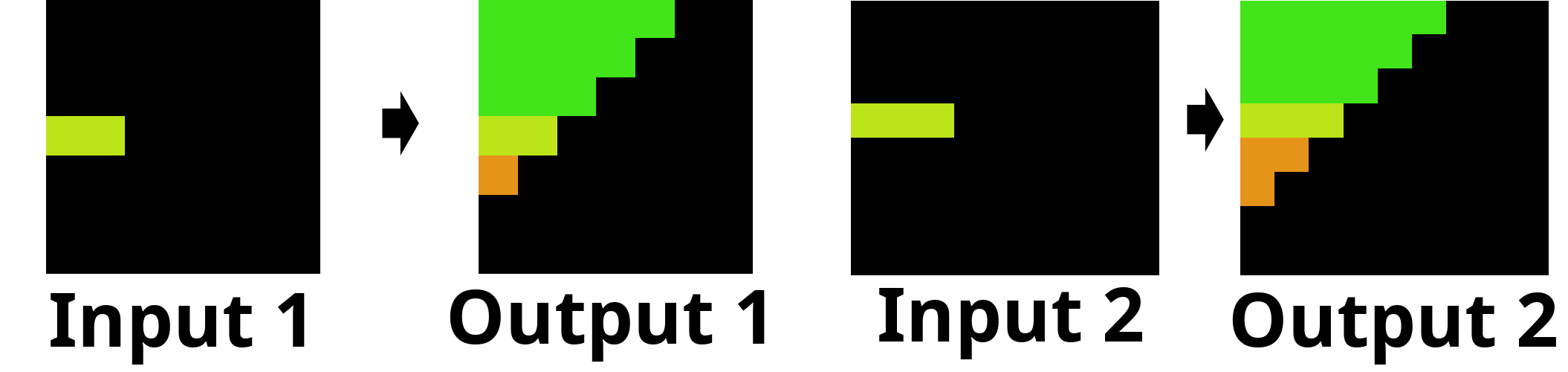}
\caption{Two training pairs used for training.}
\label{fig:problem140pairs}
\end{subfigure}

\caption{An example of solution generated by standard NCA and relative training pairs.}
\label{fig:problem140}
\end{figure}

Figure \ref{fig:NCAsol} shows an example of one of the solutions produced by the NCA model, while Figure \ref{fig:problem140pairs} shows the two training examples. In this problem, a line of a given length is presented in a random y coordinate and the correct solution corresponds to adding green lines of increasing length above the input line and orange lines of decreasing length below. Such a solution grows correctly in an incremental manner by the NCA, which generalizes to unseen y coordinates.

\begin{figure}[t!]
\centering
\begin{subfigure}[b]{0.47\textwidth}
\includegraphics[width=\textwidth]{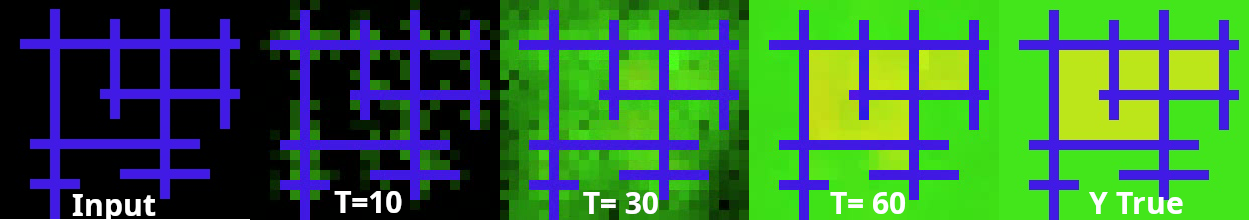}
\caption{Example solution generated by the EngramNCA v1.}
\label{fig:NgramV1sol}
\end{subfigure}

\begin{subfigure}[b]{0.47\textwidth}
\centering
\includegraphics[width=\textwidth]{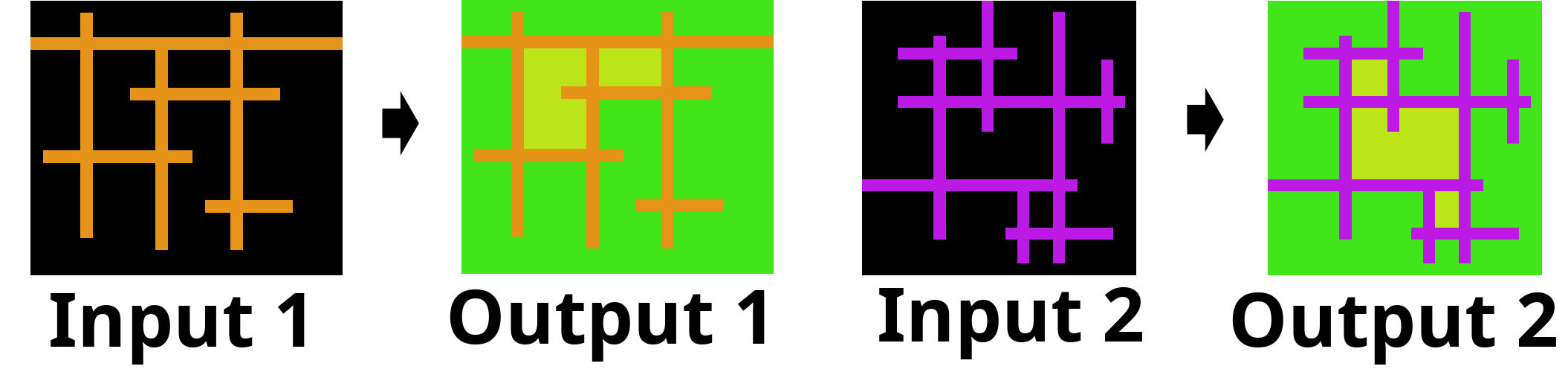}
\caption{Two training pairs used for training.}
\label{fig:problem150pairs}
\end{subfigure}

\caption{An example of solution generated by EngramNCA v3 and relative training pairs.}
\label{fig:problem150}
\end{figure}

Figure \ref{fig:problem150} shows an example of one of the solutions produced by the EngramNCA v1 model, the standard version of EngramNCA. This problem presents horizontal and vertical lines (of different color in different examples) crossing and therefore composing constrained spaces in the middle and open spaces on the outside. The correct solution fills the closed and open parts with given colors. The CA solution grows cells of green color to fill the entire space; however when they are surrounded by boundaries they are able to change to the right color. 

\begin{figure}[t!]
\centering
\begin{subfigure}[b]{0.47\textwidth}
\includegraphics[width=\textwidth]{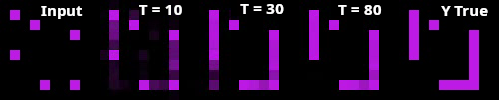}
\caption{Example  solution generated by the EngramNCA v3.}
\label{fig:NgramV3sol}
\end{subfigure}

\begin{subfigure}[b]{0.47\textwidth}
\centering
\includegraphics[width=\textwidth]{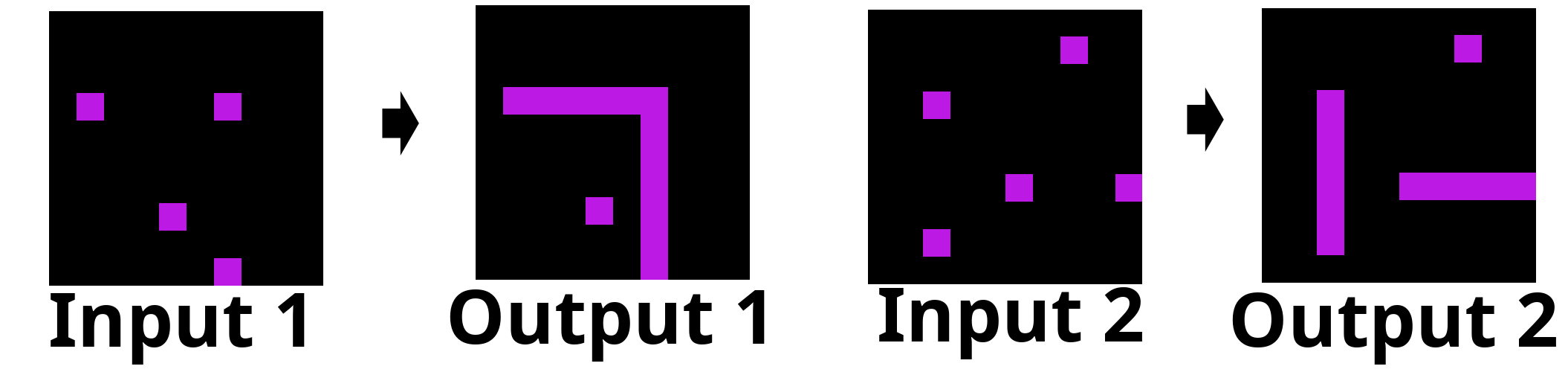}
\caption{Two training pairs used for training.}
\label{fig:problem83pairs}
\end{subfigure}

\caption{An example of solution generated by EngramNCA v3 and relative training pairs.}
\label{fig:problem83}
\end{figure}

Figure \ref{fig:problem83} shows an example of one of the solutions produced by the EngramNCA v3 model. In this test, the input contains single pixels and the correct solution connects those on the same horizontal or vertical line. The CA grows lines from the pixels and sometimes overshoots after the connecting pixel; however it is able to remove the parts not needed that reach the boundaries. 

\begin{figure}[t!]
\centering
\begin{subfigure}[b]{0.47\textwidth}
\includegraphics[width=\textwidth]{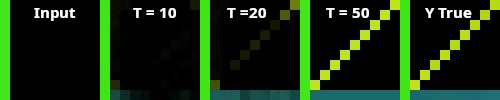}
\caption{Example solution generated by the EngramNCA v4.}
\label{fig:NgramV4sol}
\end{subfigure}

\begin{subfigure}[b]{0.47\textwidth}
\centering
\includegraphics[width=\textwidth]{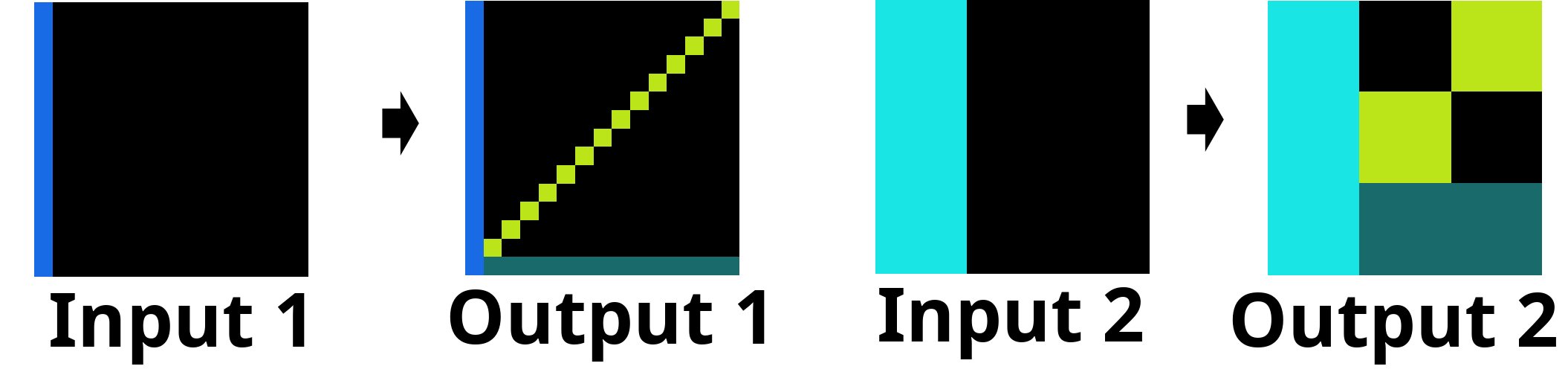}
\caption{Two training pairs used for training.}
\label{fig:program155pairs}
\end{subfigure}

\caption{An example of solution generated by EngramNCA v4 and relative training pairs.}
\label{fig:problem155}

\end{figure}

Figure \ref{fig:problem155} shows an example of one of the solutions produced by the EngramNCA v4 model. This test contains a single vertical line on the left side of the grid. The correct solution grows a horizontal line on the bottom and a diagonal line from the bottom left corner to the top right corner. The CA grows a solution that crosses the toroidal boundary and grows from both corners which eventually connects in the middle. Solutions generalize to different grid sizes.

\subsection{Almost Solved Problems }
ARC-NCA have the ability to produce partial solutions, or "almost solved" problems. These solutions typically have a few pixels wrong (or slightly wrong) but could serve as the basis for further refinement. It is also possible that these few mistakes can be removed with improvements to the architecture or simply by increasing the size of the NCA. To determine what sort of performance we would obtain by focusing on the partial solutions, we loosen the loss threshold to $-6$. Table \ref{tab:GeneralResultsloose} shows the solve rate when the loss threshold is loosened. The models solve anywhere from 2\% - 6\% more of the problems, indicating that there is potential for much better performance from relatively small adjustments. Table \ref{tab:UnionResultsLoose} indicate the unions of the results from different CA models. 

\begin{table}[h!]
\centering
\begin{tabular}{ |p{0.2\textwidth}||p{0.2\textwidth}|}
 \hline
 \multicolumn{2}{|c|}{CA Results with Loosened Threshold} \\
 \hline
 Model& Solve Rate \\
 \hline
 NCA   & 15.6\%\\
 EngramNCA v1&  \cellcolor{red!25}9.9\% \\
 EngramNCA v2& 11.8\% \\
 EngramNCA v3 & 16.4\%\\
 EngramNCA v4  & \cellcolor{green!25} 16.8\%\\
 
 \hline
 
\end{tabular}
\caption{Solve rate for CA modes when loss threshold is loosened to -6}
 \label{tab:GeneralResultsloose}
\end{table}

\begin{table}[h!]
\centering

\begin{tabular}{ |p{0.13\textwidth}||p{0.13\textwidth}|p{0.13\textwidth}|}
 \hline
 \multicolumn{3}{|c|}{CA Union Results} \\
 \hline
 CA& Mean $log(loss)$ &Solve Rate \\
 \hline
 NCA $\cup$ v1  & -3.97    & \cellcolor{red!25}18.3\%\\
 
 NCA $\cup$ v3& \cellcolor{green!25}  -4.32  & 20.2\% \\
 NCA $\cup$ v4 & -4.25 & 20.2\%\\
 v1 $\cup$ v3   &-3.98 & 18.7\%\\
 v3 $\cup$ v4   &-4.27 & \cellcolor{green!25}20.9\%\\
 v1 $\cup$ v4   &\cellcolor{red!25}-3.92 & 19.8\%\\
NCA $\cup$ v1 $\cup$ v3 $\cup$ v4& \cellcolor{gray!25}  -4.12  & \cellcolor{gray!25} 24\% \\
 
 \hline
\end{tabular}
\caption{Mean $log(loss)$ and solve rate for all six CA union variations with loosened loss threshold.}
 \label{tab:UnionResultsLoose}
\end{table}

\begin{figure}[h]
\centering
    \begin{subfigure}[b]{0.15\textwidth}
        \centering
         \includegraphics[width=\textwidth]{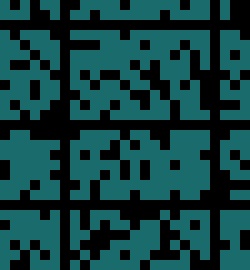}
         \caption{\footnotesize Input}
         \label{s1}
    \end{subfigure}
    \begin{subfigure}[b]{0.15\textwidth}
        \centering
         \includegraphics[width=\textwidth]{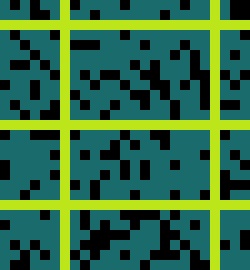}
         \caption{\footnotesize True Solution}
         \label{E1}
    \end{subfigure}
    \begin{subfigure}[b]{0.15\textwidth}
        \centering
         \includegraphics[width=\textwidth]{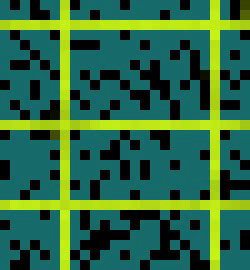}
         \caption{\footnotesize EngramNCA v3}
         \label{solv3}
    \end{subfigure}

    \begin{subfigure}[b]{0.47\textwidth}
        \centering
         \includegraphics[width=\textwidth]{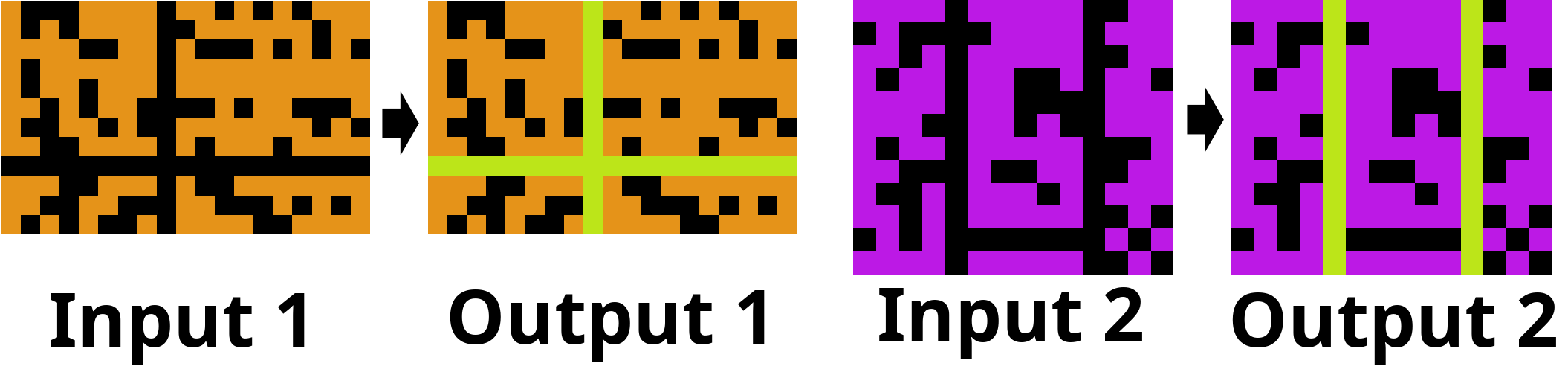}
         \caption{\footnotesize Two training pairs used for training.}
         \label{problem0pairs}
    \end{subfigure}

\caption{ An example of a near solution produced by EngramNCA v3}
\label{fig:almostsolvedv3}
\end{figure}

\begin{figure}[h]
\centering
    \begin{subfigure}[b]{0.15\textwidth}
        \centering
         \includegraphics[width=\textwidth]{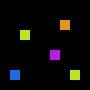}
         \caption{\footnotesize Input}
         \label{s2}
    \end{subfigure}
    \begin{subfigure}[b]{0.15\textwidth}
        \centering
         \includegraphics[width=\textwidth]{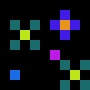}
         \caption{\footnotesize True Solution}
         \label{e2}
    \end{subfigure}
    \begin{subfigure}[b]{0.15\textwidth}
        \centering
         \includegraphics[width=\textwidth]{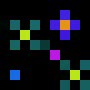}
         \caption{\footnotesize EngramNCA v1}
         \label{solv1}
    \end{subfigure}

    \begin{subfigure}[b]{0.47\textwidth}
        \centering
         \includegraphics[width=\textwidth]{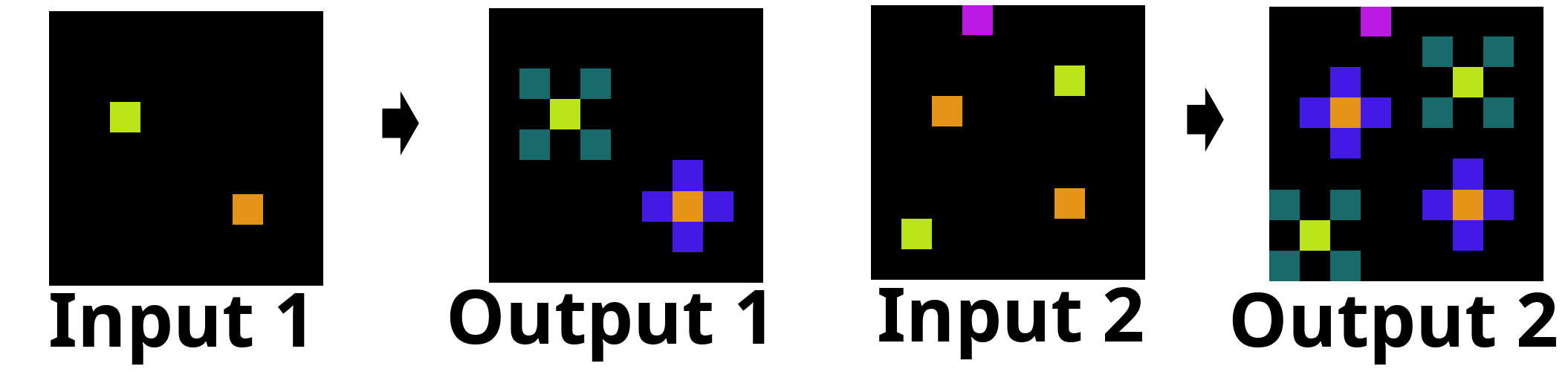}
         \caption{\footnotesize Two training pairs used for training.}
         \label{problem3pairs}
    \end{subfigure}
    
\caption{\footnotesize An example of a near solution produced by EngramNCA v1}
\label{fig:almostsolvev1}
\end{figure}

We further analyze some examples from those with minor mistakes next. Figure \ref{fig:almostsolvedv3} shows an example of a near solution produced by EngramNCA v3. We can see that the model has the general concepts to solve the problem correctly. However, three pixels are miscolored in regions with much open space. This indicates an edge case that was probably absent in the training set. Figure \ref{fig:almostsolvev1} shows an example of a near solution produced by EngramNCa v1. In this example, the model produces an exact solution at some point. However, due to the general asynchronous nature of NCA, we let the model run until it ends in a stable state. This stable state is off by one pixel.

\subsection{Reasoning Pitfalls}
Occasionally, we observe problems where the models (qualitatively) manages \textit{some} of the reasoning steps necessary to solve a particular problem, but fall short of a perfect completion. In this section we showcase some of the model-problem pair and attempt to reason about what reasoning pitfalls they might have encountered.

\begin{figure}[h]
\centering
    \begin{subfigure}[b]{0.15\textwidth}
        \centering
         \includegraphics[width=\textwidth]{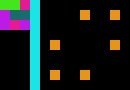}
         \caption{\footnotesize Input}
         \label{Div 1}
    \end{subfigure}
    \begin{subfigure}[b]{0.15\textwidth}
        \centering
         \includegraphics[width=\textwidth]{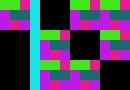}
         \caption{\footnotesize True Solution}
         \label{Div 2}
    \end{subfigure}
    \begin{subfigure}[b]{0.15\textwidth}
        \centering
         \includegraphics[width=\textwidth]{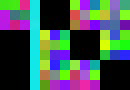}
         \caption{\footnotesize EngramNCA v4}
         \label{Div 3}
    \end{subfigure}

    \begin{subfigure}[b]{0.47\textwidth}
        \centering
         \includegraphics[width=\textwidth]{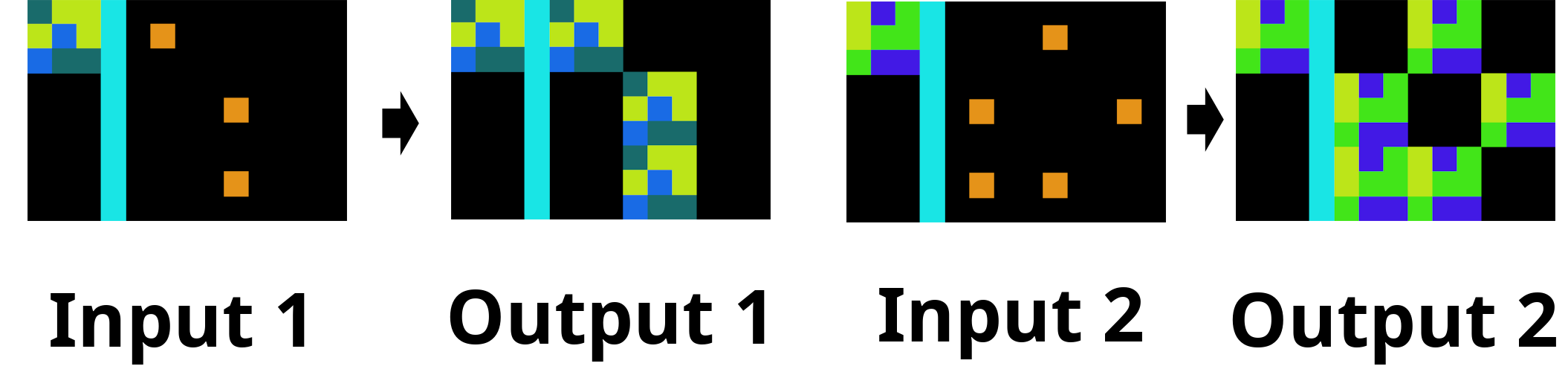}
         \caption{\footnotesize Two training pairs used for training.}
         \label{problem2pairs}
    \end{subfigure}

\caption{\footnotesize An example of partial reasoning success in a solution of ARC generated by EngramNCA v4}
\label{fig:ReasoningError}
\end{figure}

Figure \ref{fig:ReasoningError} depicts an example of a partial reasoned solution produced by the EngramNCA v4 model. 
Here we can see the model learns one of the two reasoning steps, that of growing \textit{a} pattern of the correct shape on the orange dots. However, it fails to generalise to any pattern on the left and gets the exact pixel colors wrong.

\subsection{Further Experiments}\label{sec:Further}
In this section, we detail the results of two further experiments conducted: Increasing the dimension of the hidden layer of EngramNCA v3, and solving all ARC-AGI problems by use of maximal padding as described in \nameref{sec:sizes}.

\begin{table}[h!]
\centering

\begin{tabular}{ |p{0.13\textwidth}|p{0.13\textwidth}|p{0.13\textwidth}|}
 \hline
 \multicolumn{3}{|c|}{\textbf{CA Architecture Details}} \\
 \hline
 \textbf{CA type}& \textbf{Augmentations} & \textbf{Channels, Hidden Size} \\
 \hline
 EngramNCA v3 & Sensing + Toroidal & 50, (132,132)\\
 \hline
 EngramNCA v3 & Sensing + Toroidal + Maximal Padding & 50, (132,132)\\
 \hline
\end{tabular}
\caption{Architecture detail for of EngramNCA v3 and its maximally padded version.}
 \label{tab:CAArchi2}
\end{table}

Table \ref{tab:CAArchi2} shows the architecture details for larger EngramNCA v3 and its maximally padded version. 

\begin{table}[h!]
\centering
\begin{tabular}{ |p{0.1\textwidth}||p{0.08\textwidth}|p{0.08\textwidth}|p{0.08\textwidth}|}
 \hline
 \multicolumn{4}{|c|}{CA Results} \\
 \hline
 Model & Solve Rate @ -7 & Solve Rate @-6  & Cost (\$/Task)\\
 \hline
 EngramNCA v3 & \cellcolor{green!25}16.1\% & 19.8\% &$\approx$\cellcolor{green!25} 5$e$-4\\
 EngramNCA v3 Padded   &16\% & \cellcolor{green!25}27\%&$\approx$ 7$e$-4\\
 Chat GPT 4.5$\star$ & \cellcolor{red!25}10.3\% &\cellcolor{red!25} 10.3\% &\cellcolor{red!25}0.29\\
 \hline
 
\end{tabular}
\caption{Solve rate for larger EngramNCA v3 and its maximally padded version compared to Chat GPT 4.5.
$\star$ For the sake of neatness, Chat GPT 4.5 results are displayed on the same table, even though they are not comparable through Mean $log(loss)$.}
 \label{tab:ExtraResults}
\end{table}

Table \ref{tab:ExtraResults} shows the results of EngramNCA v3 and its maximally padded version as compared to Chat GPT 4.5. By increasing the hidden size, we can observe an increase in the number of problems solved. Maximal padding increases the number of problems the CA has to solve, yet we do not see a decrease in the percentage of problems the CA can solve, suggesting that self-size modification is trivial for the CA or that the extra information provided by the padding tokens has helped with some of the problems. The maximal padding does incur a cost as NCA memory usage and run time scale poorly with lattice size. Despite this, they both still outperform Chat GPT 4.5. Leaving room for partial solutions, we see that the maximally padded version sees a significant increase in its solve rate (27\% versus 16\%).

\section{Summary and Discussion}

This work introduces ARC-NCA, a developmental framework utilizing Neural Cellular Automata to address the challenges posed by the Abstraction and Reasoning Corpus benchmark, which requires robust abstraction and reasoning capabilities derived from minimal training data. Our ARC-NCA models exploit the intrinsic properties of NCAs to emulate complex, emergent behaviors reminiscent of biological developmental processes. We evaluated standard NCA alongside several modified versions of EngramNCA, which were augmented to better accommodate specific characteristics of ARC tasks. These modifications encompassed enhanced sensing mechanisms, adjustments in local versus global information processing, and strategies for managing toroidal lattice behaviors.

The results demonstrated that ARC-NCA models achieved solve rates comparable to, and sometimes surpassing, those of popular LLMs such as ChatGPT 4.5, notably at significantly reduced computational costs. When considering partially correct solutions, success rates increased remarkably, indicating potential for further enhancements such as architectural modifications and parameters scaling. Analysis of solved and partially solved problems provided insights into the developmental nature of NCAs, revealing strengths in iterative refinement and emergent reasoning capabilities. Conversely, examples of reasoning pitfalls highlighted specific limitations in NCAs' generalization capacities, particularly in handling fine-grained details or novel edge cases not well represented in training examples.

In light of the recent introduction of ARC-AGI-2 \citep{ARCAGI2}, which presents a more challenging benchmark designed to assess AI systems' adaptability and efficiency in acquiring new skills, including symbolic interpretation, compositional reasoning, and contextual rule application, our findings hold particular relevance. ARC-AGI-2 tasks have been solved by humans in under two attempts, yet current AI systems struggle with single-digit success rates. The developmental approach proposed by ARC-NCA may provide an innovative perspective to tackling abstraction and reasoning in AI systems through developmental processes governed by local interactions, or in combination with LLMs. We therefore encourage the artificial life community to tackle open problems in artificial intelligence.

\section{Future Works}

Besides ARC-AGI-2 as a natural follow up, we outline here several research directions that warrant further investigation. 

A pre-training mechanism that could facilitate learning each single problem from the few available examples would be beneficial. Such pre-training mechanism should provide knowledge at an abstraction level that is appropriate for the type of visual reasoning required for ARC, such as basic transformations that can generalize across tasks followed by task-specific fine-tuning. Alternatively, a criticality pre-training could be an interesting direction. Criticality is a behavioral regime that is know to be ideal for different kinds of computation. One hypothesis is that NCAs at criticality would be better suited for learning ARC tasks than randomly initialized NCAs.

Our results are documented on single trials, as ARC allow submission of only two candidate solutions. However, for the sake of a more rigorous investigation, multiple runs and their stability should be investigated further. Additionally, in order to compete in the official ARC-AGI leaderboard, solutions would have to be submitted for the semi-private and private evaluation sets.

Future directions at the intersection of NCAs and LLMs are considered promising avenues. For example, LLMs may be used to suggest optimized NCA architectural choices and hyperparameters. Further, LLMs with reasoning abilities may be used as error correction mechanisms for the (almost correct) developmental solutions provided by NCAs.
Other correction mechanisms may also be considered, for example relying on NCAs or other computer vision techniques. 

Finally, NCAs operating at an abstract, latent representation \citep{menta2024latent}, may be able to capture basic primitives beneficial for reasoning, by shifting the computation from the input space to the latent space. This may be particularly relevant for architectures as EngramNCA, which try to capture basic primitives first, and then regulation mechanisms for their activation and communication.

\section{Acknowledgments}
This work was partly funded by the priority area "The Digital Society" at Østfold University College.

\footnotesize
\bibliographystyle{apalike}
\bibliography{main} 



\end{document}